\documentclass[10pt,twocolumn,letterpaper]{article}

\usepackage{iccv}
\usepackage{times}
\usepackage{epsfig}
\usepackage{graphicx}
\usepackage{amsmath}
\usepackage{amssymb}
\usepackage{algorithm}
\usepackage{algpseudocode}
\usepackage{adjustbox}
\usepackage{multirow}

\usepackage[pagebackref=true,breaklinks=true,letterpaper=true,colorlinks,bookmarks=false]{hyperref}

\iccvfinalcopy 

\def\iccvPaperID{0000} 

\ificcvfinal\fi

\def\convnet{convolutional network}
\DeclareMathOperator*{\argmax}{arg\,max}

\begin{document}

\title{STNet: Selective Tuning of Convolutional Networks for \\ Object Localization}

\author{Mahdi Biparva, John Tsotsos\\
	Department of Electrical Engineering and Computer Science\\
	York University\\
	Toronto, ON, Canada, M3J 1P3\\
	{\tt\small {mhdbprv, tsotsos}@cse.yorku.ca}}

\maketitle

\begin{abstract}
Visual attention modeling has recently gained momentum in developing visual hierarchies provided by Convolutional Neural Networks. 
Despite recent successes of feedforward processing on the abstraction of concepts form raw images, the inherent nature of feedback processing has remained computationally controversial. 
Inspired by the computational models of covert visual attention, we propose the Selective Tuning of Convolutional Networks (STNet). It is composed of both streams of Bottom-Up and Top-Down information processing to selectively tune the visual representation of \convnet s. 
We experimentally evaluate the performance of STNet for the weakly-supervised localization task on the ImageNet benchmark dataset. We demonstrate that STNet not only successfully surpasses the state-of-the-art results but also generates attention-driven class hypothesis maps. 
\end{abstract}

\section{Introduction}
\label{sec:int}
Inspired by physiological and psychophysical findings, many attempts have been made to understand how the visual cortex processes information throughout the visual hierarchy \cite{deyoe1988concurrent,kravitz2013ventral}. 
It is significantly supported by reliable evidence \cite{gilbert2013top, herzog2014vision} that information is processed in both directions throughout the visual hierarchy: the data-driven Bottom-Up (BU) processing stream convolves the input data using some forms of information transformation. In other words, the BU processing stream shapes the visual representation of the input data via a hierarchical cascading stages of information processing. On the other hand, the task-driven Top-down (TD) processing stream is perceived to modulate the visual representation such that the task requirements are completely fulfilled. 
Consequently, TD processing stream plays the role of projecting the task knowledge over the formed visual representation to achieve task requirements.

In recent years, while the learning approaches have been getting matured, various models and algorithms have been developed to present a richer visual representation for various visual tasks such as object classification and detection, semantic segmentation, action recognition, and scene understanding \cite{andreopoulos201350, dick2009evolut}. Regardless of the algorithms used for representation learning, most attempts benefit from BU processing paradigm, while TD processing has very rarely been targeted particularly in the computer vision community. In recent years, \convnet s, as a BU processing structure, have shown to be quantitatively very successful on the visual tasks targeted by popular benchmark datasets  \cite{krizhevsky2012imagenet,szegedy2014going,HeZhangRenEtAl2015,girshick2015fast}.

\begin{figure}[t]	
	\begin{center}	
		\includegraphics[width=1.0\columnwidth]{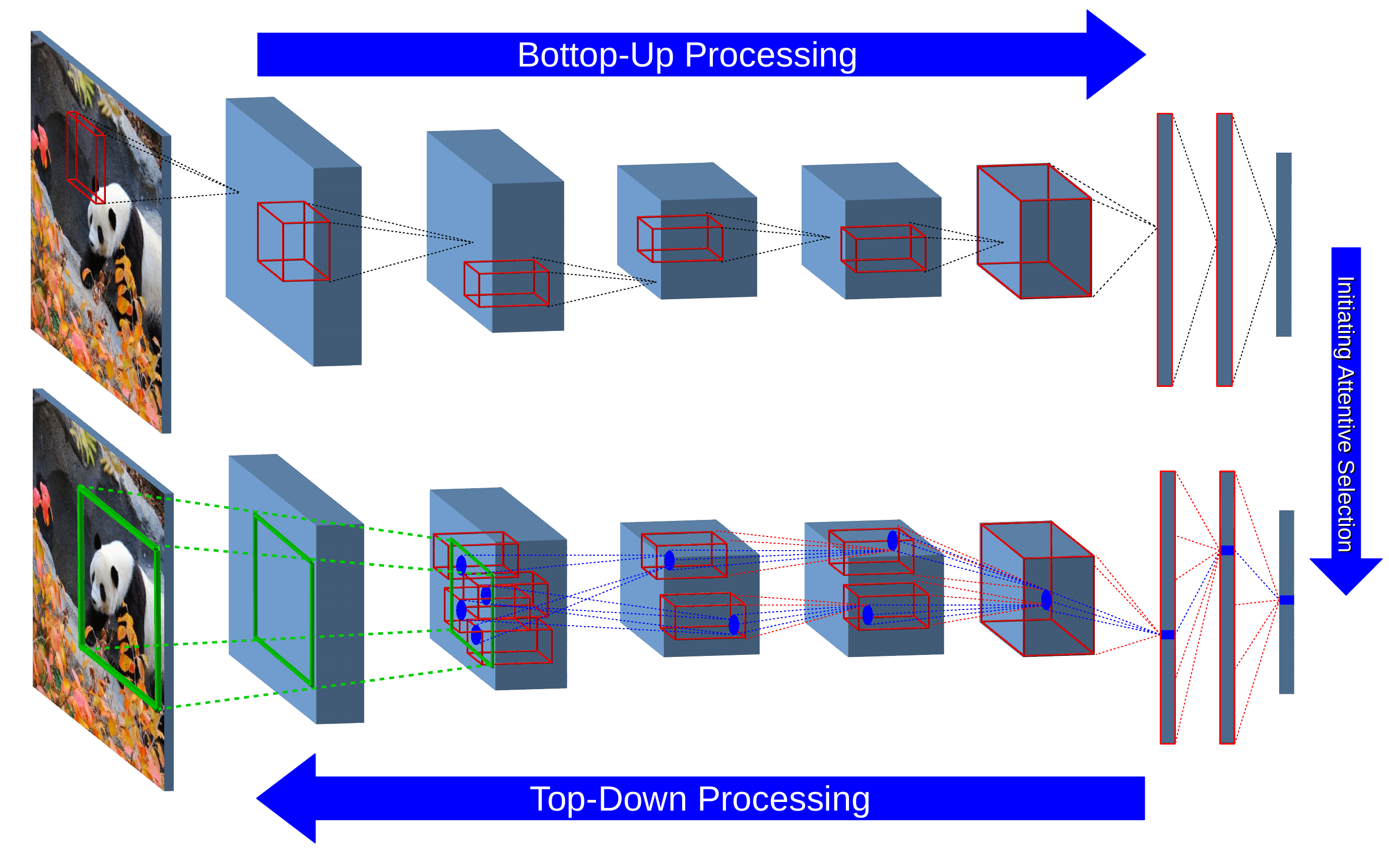}	
	\end{center}	
	\caption{STNet consists of both BU and TD processing streams. In the BU stream, features are collectively extracted and transferred to the top of the hierarchy at which label prediction is generated. The TD processing, on the other hand, selectively activate part of the structure using attention processes.}	
	\label{fig:stnet}
\end{figure}

Attempts in modeling visual attention are attributed to the TD processing paradigm. The idea is using some form of facilitation or suppression, the visual representation is selected and modulated in a TD manner  \cite{tsotsos2011computational,lee2013critical}. Visual attention has two modes of execution \cite{carrasco2011visual,kowler2011eye}: Overt attention attempts to compensate for the lack of visual acuity throughout the entire field of view in a perception-cognition-action cycle by the means of an eye-fixation controller. In the nutshell, the eye movement keeps the highest visual acuity at the fixation while leaving the formed visual representation intact. Covert attention, on the other hand, modulates the shaped visual representation, while keeping the fixation point unchanged.

We strive to account for both the BU and TD processing in a novel unified framework by proposing STNet, which integrates attentive selection processes into the hierarchical representation. STNet has been experimentally evaluated on the task of object localization. Unlike all previous approaches, STNet considers the biologically-inspired method of surround suppression \cite{tsotsos1990analyzing} to selectively deploy high-level task-driven attention signals all the way down to the early layers of the visual hierarchy. The qualitative results reveal the superiority of STNet on this task over the performance of the state-of-the-art baselines. 

The remaining of the paper is organized as follows. In Section \ref{sec:int}, we review the related work on visual attention modeling in the computer vision community. Section \ref{sec:mod} presents the proposed STNet model in details. Experiments are conducted in Section \ref{sec:exp} in which STNet performance is qualitatively and quantitatively evaluated. Finally, the paper is ended by a conclusion in Section \ref{sec:con}.

\section{Related Work}
\label{sec:rel}

In recent years, the computer vision community has gained momentum in improving the evaluation results on various visual tasks by developing various generations of deep learning models. \convnet s have shown their superiority in terms of the learned representation for tasks inherently related to visual recognition such as object classification and detection, semantic segmentation, pose estimation, and action recognition. 

Among various visual attention models \cite{bylinskii2015towards}, covert visual attention paradigm involves the scenario in which eye movement is not considered throughout the modeling approach.
Fukushima's attentive Neocognitron \cite{fukushima1986neural} proposes that attention can be seen as a principled way to recall noisy, occluded, and missing parts of an input image. TD processing is realized as a form of facilitatory gain modulation of hidden nodes of the BU processing structure. 
Selective Tunning Model of visual attention \cite{tsotsos1995SelTun}, on the other hand, proposes TD processing using two stages of competition in order to suppress the irrelevant portion of the receptive field of each node. Weights of the TD connections, therefore, are determined as the TD processing continues. Furthermore, only the BU hidden nodes falling on the trace of TD processing are modulated while leaving all the rest intact.

Various attempts have been made to model an implicit form of covert attention on \convnet s. \cite{simonyan2013deep} proposes to maximize the class score over the input image using the backpropagation algorithm for the visualization purposes. \cite{zeiler2014visualizing} introduces an inversed \convnet to propagate backward hidden activities to the early layers. Harnessing the superiority of global \textit{AVERAGE} pooling over global \textit{MAX} pooling to preserve spatial correlation, \cite{zhou2016learning} has defined a weighted sum of the activities of the convolutional layer feeding into the global pooling layer.
Recently, an explicit notion of covert visual attention has gained interest in the computer vision community \cite{cao2015look, zhang2016top} for the weakly-supervised localization task. Having interpreted \textit{ReLU} activation and \textit{MAX} pooling layers as feedforward control gates, \cite{cao2015look} proposes feedback control gate layers which are activated based on the solution of an optimization problem. Inspired closely by Selective Tunning model of visual attention, \cite{zhang2016top} formulates TD processing using a probabilistic interpretation of the \textit{Winner-Take-All} (WTA) mechanism. In contrast to all these attempts that the TD processing is densely deployed in the same fashion as BU processing, we propose a highly sparse and selective TD processing in this work.

The localization approach in which the learned representation of the visual hierarchy is not modified is commonly referred to as weakly supervised object localization \cite{simonyan2013deep, oquab2015object, zhou2016learning, cao2015look, zhang2016top}. This is in contrast with the supervised localization approach in which the visual representation is fine-tunned to better cope with the new task requirements. Additionally, Unlike the formulation for the semantic segmentation task \cite{pathak2015constrained,papandreou2015weakly,pinheiro2015image}, bounding box prediction forms the basis of performance measure.
We evaluate experimentally STNet in this paradigm and provide the evidence that selective tuning of \convnet s better addresses object localization in the weakly-supervised regime.

\section{Model}
\label{sec:mod}
\subsection{STNet}
An integration of the conventional bottom-up processing by \convnet s with the biologically-plausible attentive top-down processing in a unified model is proposed in this work. STNet consists of two interactive streams of processing: The BU stream has the role of forming the representation throughout the entire visual hierarchy. Apparently, information is very densely processed layer by layer in a strict parallel paradigm. The BU pathway processes information at each layer using a combination of basic operations such as convolution, pooling, activation, and normalization functions. The TD stream, On the other hand, develops a projection of the task knowledge onto the formed hierarchical representation until the task requirements are fulfilled. Depending on the type of the task knowledge, the projections may be realized computationally using some primitive stages of attention processing. The cascade flow of information throughout both streams is layer by layer such that once information at a layer is processed, the layer output is fed into the next adjacent layer as the input according to the hierarchical structure.

Any computational formulation of the visual hierarchy representing the input data can be utilized as the structure of the BU processing stream as long as the primary visual task could be accomplished. Convolutional neural networks trained in the fully supervised regime for the primary task of object classification are mainly focused in this paper. 
Having STNet composed of a total of $L$ layers, the BU processing structure is composed of $\forall l \in \{0, \dots, L\}, \exists\, \mathbf{z}^l \in \mathbb{R}^{H^l \times W^l \times C^l}$, where $\mathbf{z}^l$ is the three dimensional feature volume of hidden nodes at layer $l$ with the dimension of width $W^l$, height $H^l$ and $C^l$ number of channels.

\subsection{Structure of the Top-Down Processing}
Based on the topology and connectivity of the BU processing stream, an interactive structure for the attentive TD processing is defined. According to the task knowledge, the TD processing stream is initiated and consecutively traversed downward layer by layer until the layer that satisfies task requirements is reached.
A new type of nodes is defined to interact with the hidden nodes of the BU processing structure. According to the TD structure, gating nodes are proposed to collectively determine the TD information flow throughout the visual hierarchy. Furthermore, they are very sparsely active since the TD processing is tuned to activate relevant parts of the representation. 

The TD processing structure consists of $\forall l \in \{0, \dots, L\}, \exists\: \mathbf{g}^l \in \mathbb{R}^{H^l \times W^l \times C^l}$, where $\mathbf{g}^l$ is the three dimensional (3D) gating volume at layer $i$ having the exact size of it's hidden feature volume counterpart in the BU processing structure. 
We define function $RF(z)$ to return the set of all the nodes in the layer below that falls inside the receptive field of the top node according to the connectivity topology of the BU processing structure.

Having defined the structural connectivity of both the BU and TD processing streams, we now introduce the attention procedure that locally processes information to determine connection weights of the TD processing structure and consequently the gating node activities at each layer. 
Once the information flow in the BU processing stream reaches the top of the hierarchy at layer $L$, the TD processing is initiated by setting the gating node activities of the top layer as illustrated in Fog. \ref{fig:stnet}. Weights of the connections between the top gating node $g_L$ and all the gating node in the layer below within the $RF(g_L)$ are computed using the attentive selection process. Finally, the gating node activities of layer $L-1$ are determined according to the connection weights. This attention procedure is consecutively executed layer by layer downward to a layer the task requirements are fulfilled.

\subsection{Stages of Attentive Selection}
Weights of the connections of the BU processing structure are learned by the backpropagation algorithm \cite{rumelhart1985learning} in the training phase.
For the TD processing structure, however, weights are computed in an immediate manner using the deterministic and procedural selection process from the Post-Synaptic (PS) activities. We define $\forall\: g^{l}_{w,h,c} \in\, \mathbf{g_{l}}\;,\: PS(g^{l}_{w,h,c}) = RF(z^{l}_{w,h,c}) \odot\: k^{l}_{c} $, where $PS(g)$ is the dot product of two similar-size matrices, one representing the receptive field activities, and the other, the kernel at channel $c$ and layer $l$.

\begin{figure}[t]	
	\begin{center}	
		\includegraphics[width=0.5\textwidth]{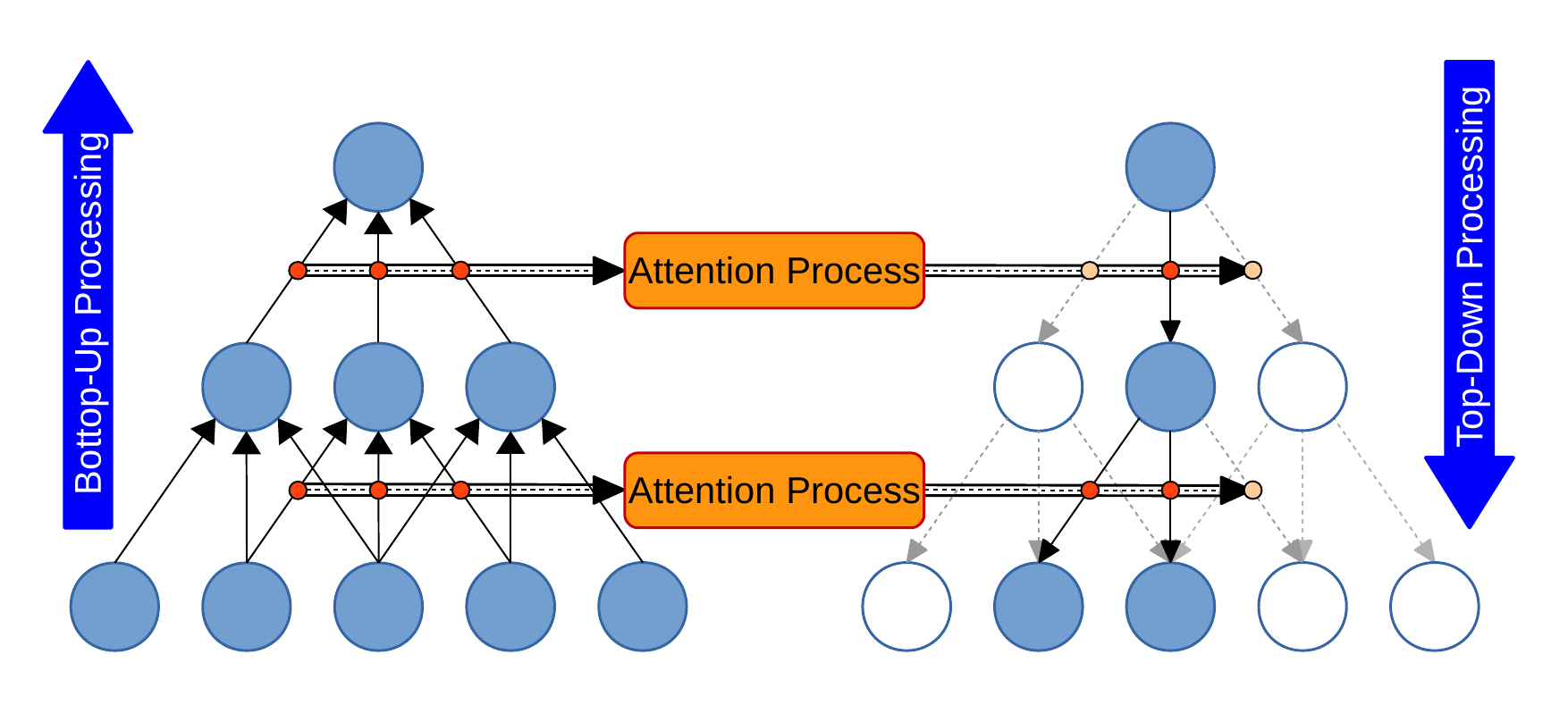}	
	\end{center}	
	\caption{Schematic Illustration of the sequence of interaction between the BU and TD processing. Three stages of the attentive selection process are illustrated.} 
	\label{fig:flow}
\end{figure}

The selection process has three stages of computation. Each stage processes the input PS activities and then feed the selected activities to the next stage. In the first stage, noisy redundant activities that interfere with the definition of task knowledge are determined to prune away. Among the remained PS activities, the most informative group of activities are marked as the winners of the selection process at the end of the second stage. In the last stage, the winner activities are normalized. Once multiplicatively biased by the top gating node activity, the activity of the bottom gating node is updated consequently. Fig. \ref{fig:flow} schematically illustrates the sequence of actions beginning from fetching PS activities from the BU stream to propagating weighted activities of the top gating node to the lower layer.

\textbf{Stage 1: Interference Reduction} 

The main critical issue to accomplish successfully any visual task is to be able to distinguish relevant regions from irrelevant ones. Winner-Take-All (WTA) is a biologically-plausible mechanism that implements a competition between input activities. At the end of the competition, the winner retains it's activity, while the rest become inactive. The Parametric WTA (P-WTA) using the parameter $\theta$ is defined as $P\text{-}WTA(PS(g),\: \theta) = \{ s\: |\: s \in\: PS(g),\: s \geq WTA(PS(g)) - \theta \}$. The role of the parameter $\theta$ is to establish a safe margin from the winner activity to avoid under-selection such that multiple winners will be selected at the end of the competition. It is remarkably critical to have some near optimal selection process at each stage to prevent the under- or over-selection extreme cases.

\begin{figure}[t]	
	\begin{center}
		\includegraphics[width=0.5\textwidth]{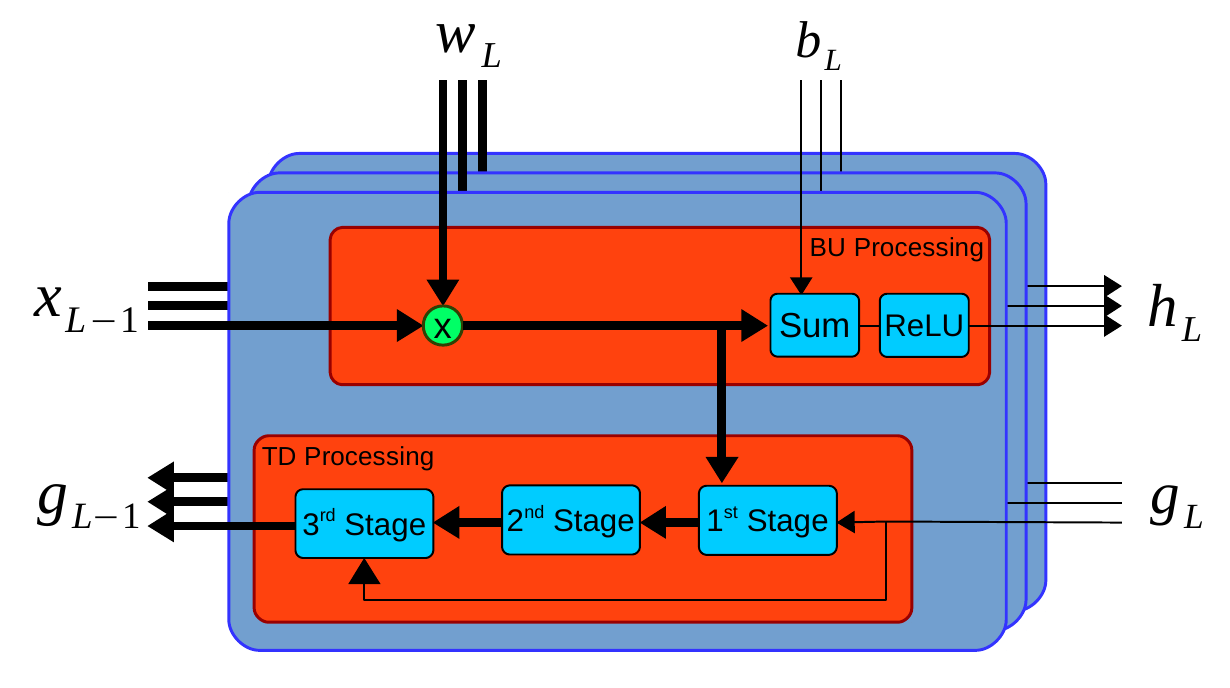}	
	\end{center}	
	\caption{Modular diagram of the interactions between various blocks of processing in both the BU and TD streams.}	
	\label{fig:modular}
\end{figure}

We propose an algorithm to tune the parameter $\theta$ for an optimal value at which the safe margin is defined based a biologically-inspired approach. It is biologically motivated that once the visual attention is deployed downward to a part of the formed visual hierarchy, those nodes falling on the attention trace will eventually retain their node activities regardless of the intrinsic selective nature of attention mechanisms \cite{reynolds1999role,tsotsos2011computational}.
In analogy to this biological finding, the Activity Preserve (AP) algorithm optimizes for the distance from the sole winner of the WTA algorithm at which if all the PS activities outside the margin are pruned away, the top hidden node activity will be preserved.


Algorithm \ref{alg:stage1} specifies the upper and lower bounds of the safe margin. The upper bound is clearly indicated by the sole winner given by the WTA algorithm, while the lower bound is achieved by the output of the AR algorithm.
Consequently, the P-WTA algorithm returns all the PS activities that fall within this range specified by the upper and lower bound values. They are highlighted as the winners of the first stage of the attentive selection process.
Basically, the set $W^{1st}$, returned from P-WTA algorithm, contains those nodes within the receptive field that most participate in the calculation of the top node activity. Therefore, they are the best candidates to initiate the attention selection processes of the layer below. The size of the set of winners at this point, however, is still large. Apparently, further stages of selection are required to prohibit interference and redundant TD processing caused by the over-selection phenomenon.

\begin{algorithm}
	\small
	\caption{Parametric $WTA$ Optimization}
	\label{alg:stage1}
	\begin{algorithmic}[1]
		\State $NEG(PS) = \{ s\: |\: s \in\: PS(g), s \le 0 \}$
		\State $POS(PS) = \{ s\: |\: s \in\: PS(g), s > 0 \}$
		\State $SUM(NEG)= \sum_{n_i \in NEG(PS)}n_i$
		\State $buffer = SUM(NEG)$
		\State $i = 0$
		\While {$i \le |POS(PS)|,\: buffer < \epsilon$}
		\State $buffer += SORT(POS(PS))[i]$
		\State $ i += 1$
		\EndWhile
		\State \textbf{return} $SORT(POS(PS))[i-1]$
	\end{algorithmic}
\end{algorithm}

\textbf{Stage 2: Similarity Grouping} 

In the second stage, the ultimate goal is to apply a more restrictive selection procedure in accordance with the rules elicited from the task knowledge. 
Grouping of the winners according to some similarity measures serves as the basis of the second stage of the attention selection process. Two modes of selection at the second stage are proposed depending on whether the current layer of processing has a spatial dimension or not: Spatially-Contiguous(SC) and Statistically-Important(SI) selection modes respectively. The former is applicable to the Convolutional layers and the latter to the Fully-Connected(FC) layers in a typical \convnet.

There is no ordering information between the nodes in the FC layers. Therefore, one way to formulate the relative importance between nodes is using the statistics calculated from the sample distribution of node activities. SI selection mode is proposed find the statistically important activities. Based on an underlying assumption that the node activities have a Normal distribution, the set of winners of the second stage is determined by $W^{2nd} = \{ s |\: s \in\: W^{1st}, s > \mu + \alpha * \sigma \}$, where $\mu$ and $\sigma$ are the sample mean and standard deviation of $W^{1st}$ respectively. The best value of the coefficient $\alpha$ is searched over the range $\{ -3, -2, -1, 0, +1, +2, +3 \}$ in the second stage based on a search policy meeting the following criteria: First, the size of the winner set $W^{2nd}$ at the end of the SI selection mode has to be non-zero. Second, the search iterates over the range of possible coefficient values in a descending order until $|W^{2nd}| \neq 0$. Furthermore, an offset parameter $O$ is defined to loosen the selection process at the second stage once these criteria are met. For instance, suppose $\alpha$ is +1 when the second stage search is over. The loosened coefficient $\alpha$ will be -1 for the offset value of 2. In Sec. \ref{sec:exp}, experimental evaluations demonstrate the effects of loosening CI selection mode on performance improvement.

Convolutional layers, on the other hand, benefit from stacks of two dimensional spatial feature maps. Although the ordering of feature maps in the third dimension is not meant to encode for any particular information, 2D feature maps individually highlight spatial active regions of the input domain projected into a particular feature space. In other words, the spatial ordering is always preserved throughout the hierarchical representation. With the spatial ordering  and the task requirement in mind, SC selection mode is proposed to determine the most spatially contiguous region of the winners based on their PS activities.

SC selection mode first partitions the set of winners $W^{1st}$ into groups of connected regions. A node has eight immediate adjacent neighbors. A connected region $R_i$, therefore, is defined as the set of all nodes that are recursively in the neighborhood of each other. Out of all the number of connected regions, the output of the SC selection mode is the set of nodes $W^2$ that falls inside the winner connected region. The winning region is determined by the index $i$ such that $\hat{i} = \argmax_{i} \alpha*\sum_{r_j\in R_i} PS_{r_j}(g) + (1-\alpha)*|R_i|$, where $PS_{r_j}(g)$ is the PS activity of node $r_j$ among the set of all PS activities of the top node g, and the value of multiplier $\alpha$ is cross-validated in the experimental evaluation stage for a balance between the sum of PS activities and the number of nodes in the connected region. Lastly, SC selection mode returns the final set of winners $W^{2nd} = \{ s |\: s \in\: R_{\hat{i}} \}$ such that $W^{2nd}$ better addresses task requirements in comparison to $W^{1st}$. We experimentally support this arguments in \ref{sec:exp}.

Having known the set of winners $W^{2nd}$ out of the entire set of nodes falling inside the receptive field of the top node $RF(g)$, it is straightforward to compute values of the weight connections of the TD processing structure that are active and those that remain inactive. The inactive weight connections have value zero. In Stage 3, the mechanism to set the values of the active weight connections from $W^{2nd}$ will be described.

\begin{table}[t]
	\begin{adjustbox}{width=1.0\columnwidth,center}
		\begin{tabular}{l|ccccc}
			\hline
			Architecture		& $L_{prop}$ & $O_{FC}$ & $O_{Bridge}$ & $\alpha$ & $\delta_{post}$\\ \hline
			ST-AlexNet    & pool1    & 3 & 3 & 0.2 & $\mu_A$  \\
			ST-VGGNet     & pool3 & 2 & 0 & 0.2 & $\mu_A$ \\
			ST-GoogleNet  & pool2/3x3\_s2     & 0 & - & 0.2 & $\mu_A$ \\	\hline
		\end{tabular}
	\end{adjustbox}
	\caption{Demonstration of the STNet configurations in terms of  The hyperparameter values. $L_{prop}$ is the name of the layer at which the attention map is calculated. $O_{FC}$ and $O_{Bridge}$ are the offset values of the CI selection mode at the fully-connected and bridge layers respectively. $\alpha$ is the trade-off multiplier of the SC selection mode. $\delta_{post}$ represents the post-processing threshold value of the attention map.}
	\label{table:hyparam}
\end{table}

\textbf{Stage 3: Attention Signal Propagation}

Gating node are defined to encode attention signals using multiple level of activities. The top gating node propagates the attention signal proportional to the normalized connection weights to the layer below. 
Having the set of winners $W^{2nd}$ for the top gating node $g$, $PS_{W^{2nd}}(g)$ is the set of PS activities of the corresponding winners. The set of normalized PS activities is defined as $PS_{norm}=\{ s' |\: s \in\: PS_{W^{2nd}}(g),\: s' = s / \sum_{s_i\in PS_{W^{2nd}}(g)} s_i \}$. Weight values of the active TD connections are specified as follows: $\forall\: i \in W^{2nd}, w_{ig}=PS_{norm}^i$, where $w_{ig}$ is the connection from the top gating node $g$ to the gating node $i$ in the layer below, and $PS_{norm}^i$ is the PS activity of the winner node $i$.

At each layer, the attentive selection process is performed for all the active top gating nodes. Once the winning set for each top gating node is determined and the normalized values of the corresponding connection weights to the layer below are computed, the winner gating nodes of the layer below are updated as follows: $\forall\: i \in \{1, \dots, |\textbf{g}^l|\},$ $\forall\: j \in \{1, \dots, |W^{2nd}_i|\},\: g^{l-1}_j += w_{ji}*g^{l}_i$. The updating rule ensures that the top gating node activity is propagated downward such that it is multiplicatively biased by weight values of the active connections.

\section{Experimental Results}
\label{sec:exp}
Top-down visual attention is necessary for the completion of sophisticated visual tasks for which only Bottom-Up information processing is not sufficient. This implies that tasks such as object localization, visual attribute extraction, and part decomposition require more processing time and resources. STNet, as a model benefiting from both streams of processing, is experimentally evaluated on object localization task in this work.

\begin{table}[t] 
	\begin{adjustbox}{width=0.8\columnwidth,center=0.5\textwidth}
		\begin{tabular}{l|ccc}
			\hline
			Model		& AlexNet & VGGNet & GoogleNet \\ \hline
			Oxford\cite{simonyan2013deep}      & - & - & 44.6  \\
			CAM*\cite{zhou2016learning}         & - & - & 48.1  \\			
			Feedback\cite{cao2015look}   & 49.6 & 40.2 & 38.8  \\
			MWP\cite{zhang2016top}     & - & - & 38.7  \\	\hline
			STNet		& \textbf{40.3} & \textbf{40.1} & \textbf{38.6}  \\ \hline
		\end{tabular}
	\end{adjustbox}
	\caption{Comparison of the STNet localization error rate on the ImageNet validation set with the previous state-of-the-art results. The bounding box is predicted given the single center crop of the input images with the TD processing initialized by the ground truth category label. 
		*Based on the result reported by \cite{zhang2016top}}
	\label{table:localresults}
\end{table}

STNet is implemented using Caffe \cite{jia2014caffe}, a library originally developed for \convnet s. AlexNet \cite{krizhevsky2012imagenet}, VGGNet(16) \cite{simonyan2013deep}, and GoogleNet \cite{szegedy2014going} are the three \convnet $ $ architectures that are applied to define the BU processing structure of STNet. The model weight parameters are retrieved from the publicly available \convnet $ $ repository of Caffe Model Zoo in which they are pre-trained on ImageNet 2012 classification training dataset \cite{deng2009imagenet}. For the rest of the paper, we refer to STNet with AlexNet as the base architecture of the BU structure as ST-AlexNet. This also applies similarly to VGGNet and GoogleNet.

\begin{figure*}[t]	
	\begin{center}
		\includegraphics[width=1.0\textwidth]{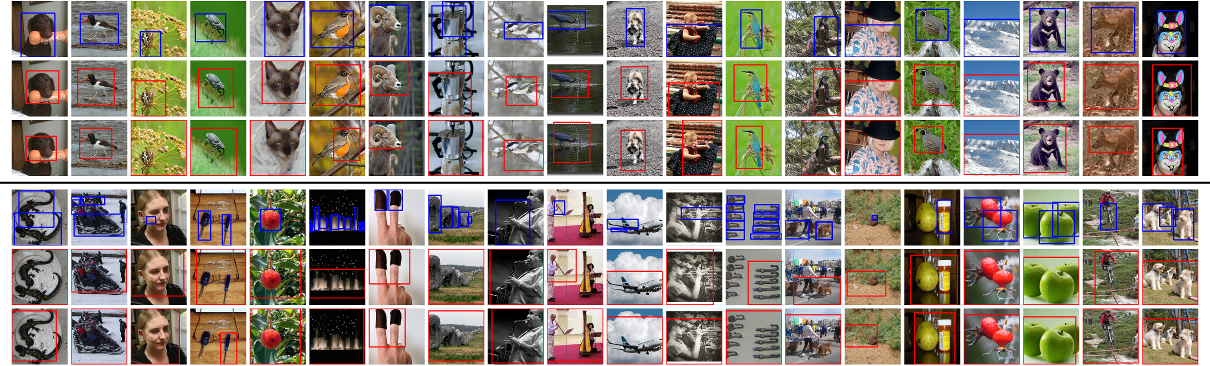}	
	\end{center}	
	\caption{Illustration of the predicted bounding boxes in comparison to the ground truth for ImageNet images. In the top section, STNet is successful to localize the ground truth objects. The bottom section, on the other hand, demonstrates the failed cases. The top, middle, and bottom rows of each section depict the bounding boxes from the ground truth, ST-VGGNet, and ST-GoogleNet respectively.}
	\label{fig:phase1}
\end{figure*}

\subsection{Implementation Details}
\label{subsec: impdetails}

\textbf{Bounding Box Proposal:}
Having an input image fed into the BU processing stream, a class specific attention map for category $k$ at layer $l$ is created. It is a resultant of the TD processing stream initiated from the top gating layer with only one at node $k$ and zero at the rest. Once the attention signals are completely propagated downward to layer $l$, the class specific attention map is defined by collapsing the gating volume $\mathbf{g}^l \in \mathbb{R}^{H^l \times W^l \times C^l}$ at the third dimension into the attention map $A^l_k \in \mathbb{R}^{H^l \times W^l}$ as follows: $A^l_k=\sum_{i \in C^l} g^l_i$, where $C^l$ is the number of gating sheets at layer $l$, and $g^l_i$ is a 2D gating sheet. We propose to post-process the attention map by setting all the small collapsed values below the sample mean value of the map to zero. 


We propose to predict a bounding box from the thresholded attention map $\hat{A}^l_k$ using the following procedure. Apparently, the predicted bounding box is supposed to enclose an instance of the category $k$. If layer $l$ is somewhere in the middle of the visual hierarchy, $\hat{A}^l_k$ is transformed into the spatial space of the input layer. In the subsequent step, a tight bounding box around the non-zero elements of the transformed $\hat{A}^l_k$ is calculated. Nodes inside the RF of the gating nodes at the boundary of the predicted box are likely to be active if the TD attentional traversal further continues processing lower layers. Therefore, we choose to pad the tight predicted bounding box with the half size of the accumulated RF at layer $l$. We calculate accurately the accumulated RF size of each layer according to the intrinsic properties of the BU processing structure such as the amount of padding and striding of the layer.


\textbf{Search over Hyperparameters:}
There a few number of hyperparameters in STNet that are experimentally cross-validated using one held-out partition of the ImageNet validation set. It contains 1000 images which are selected from the randomly-shuffled validation set. The grid search over the hyperparameter space finds the best-performing configuration for each \convnet $ $ architecture.

The SI selection mode is experimentally observed to perform more efficiently once the offset parameter $O$ is higher than zero. The offset parameter has the role of loosening the selection process for the cases under-selection is very dominant. 
Furthermore, we define the bridge layer as the one at which the 3D volume of hidden nodes collapses into a 1D hidden vector. CI selection procedure is additionally applied to the entire gating volume of the bridge layer in order to prevent the over-selection phenomenon. Except GoogleNet, the other two have a bridge layer. Further implementation details regarding all three architectures are given in the supplementary document. 

Hyperparameters such as the layer at which the best localization result is obtained, the multiplier of the SC selection mode, and the threshold value for the bounding box proposal procedure are all set by the values obtained from the cross-validation on the held-out partition set for all three \convnet s. Having the best STNet configurations given in Table \ref{table:hyparam}, we measure STNet performance on the entire ImageNet validation set.

\begin{figure*}[t]	
	\begin{center}
		\includegraphics[width=1.0\textwidth]{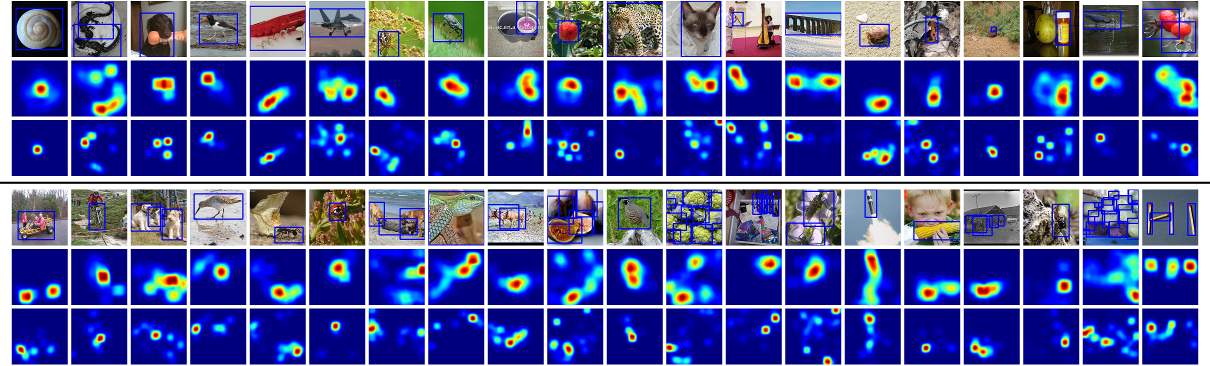}	
	\end{center}	
	\caption{Demonstration of the attention-driven class hypothesis maps for ImageNet images. In each top or bottom section, each row top to bottom represents ground truth boxes on RGB images, the CH map from ST-VGGNet, and the CH map from ST-GoogleNet respectively.}	
	\label{fig:phase2}
\end{figure*}

\subsection{Weakly Supervised Localization}
\label{sec:weak}
The significance of the attentive TD processing in STNet is both quantitatively and qualitatively evaluated on the ImageNet 2015 benchmark dataset for the object localization task. The experimental setups and procedures have been considerably kept comparative with previous works.

\textbf{Dataset and evaluation:}
Localization accuracy of STNet is evaluated on the ImageNet 2015 validation set containing 50,000 images of variable sizes. The shortest side of each image is reduced to the size of the STNet input layer. A single center crop of the size equal to the input layer is then extracted and sent to STNet for bounding box prediction. In order to remain comparative with the previous experimental setups for the weakly supervised localization task \cite{cao2015look, zhang2016top}, the ground truth label is provided to initiate the TD processing. A localization prediction considers to be correct if the Intersection-over-Union (IoU) of the predicted bounding box with the ground truth is over $0.5$.

\textbf{Quantitative results:}
STNet localization performance surpasses the previous works with a comparative testing protocol on the ImageNet dataset. For all three BU architectures, Table \ref{table:localresults} indicates that STNet quantitatively outperforms the state-of-the-art results \cite{simonyan2013deep,zhou2016learning,cao2015look,zhang2016top}. The results imply not only the localization accuracy has improved but also significantly less number of nodes are active in the TD processing stream, while all the previous approaches densely seek a locally optimum state of the TD structure.

\textbf{Comparison with Previous Works:}
One of the factors distinguishing STNet from other approaches is the selective nature of the TD processing. In gradient-based approaches such as \cite{simonyan2013deep}, the gradient signals, which are computed with respect to the input image rather than the weight parameters, are deployed densely downward to the input layer. Deconvnet \cite{zeiler2014visualizing} is proposed to reverse the same type and extent of processing as the feedforward pass originally for the purpose of visualization. The Feedback model \cite{cao2015look} similarly defines a dense feedback structure that is iteratively optimized using a secondary loss function to maintain the label predictability of the entire network. The recent MWP model \cite{zhang2016top} proposes to remain faithful to all the aforementioned models with respect to the extent of the TD processing. In contrast to these all, the TD structure of STNet remains fully inactive except to a small portion that leads to the attended region of the input image. We empirically verify that on average less than $0.3\%$ of the TD structure is active, while improving the localization accuracy.
This implies comparative localization results can be obtained with faster speed and less wasted amount of computation in the TD processing stream. Furthermore, it is worth noting that ST-AlexNet localization performance is very close to the two other high capacity models despite the shallow depth and simplicity of the network architecture.

\textbf{Qualitative Analysis:}
The qualitative results provide insights on the strengths and weakness of STNet as illustrated in Fig. \ref{fig:phase1}. 
Investigating the failed cases, we are able to identify two extreme scenarios: under-selection and over-selection scenarios. The under-selection scenario is caused by the inappropriate learned representation or improper configuration of the TD processing, while the over-selection scenario mainly is due to either multi-instance or what we call \textit{Correlated Accompanying Object} cases. A large bounding box enclosing multiple objects is proposed as a result of over-selection. Neither streams of STNet are tuned to systematically deal with these extreme scenarios. 



\subsection{Class Hypothesis Visualization}
\label{sec:saliency}
We show that gating node activities can further be processed to visualize the salient regions of input images for an activated category label.
Following a similar experimental setup to the localization task given in Table \ref{table:hyparam}, an attention-driven \textit{Class Hypothesis} (CH) map is created from the transformed thresholded attention map. We simply increment by one the pixel values inside the accumulated RF box centered at each non-zero pixel of the attention map.
Once iterated over all non-zero pixels, the CH map is smoothed out using a Gaussian filter with the standard deviation $\sigma=6$. Fig. \ref{fig:phase2} illustrates qualitatively the performance of STNet to highlight the salient parts of the input image once the TD processing stream is initiated with the ground truth category label. Further details regarding the visualization experimental setups are given in the supplementary document.

\textbf{Comparison of \convnet s}:
We observed in Sec. \ref{sec:weak} that the localization performance of the ST-GoogleNet surpasses both ST-AlexNet and ST-VGGNet. 
The qualitative experimental results using CH maps in Fig. \ref{fig:phase2} further shed some light on the inherent nature of this discrepancy.
Both AlexNet and VGGNet take benefit of a coherently increasing RF sizes along the visual hierarchy such that at each layer all hidden nodes have a similar RF size. Consequently, the scale at which features are extracted coherently changes from layer to layer. On the other hand, GoogleNet is always taking advantage of intermixed multi-scaling feature extraction at each layer. Additionally, 1x1 convolutional layers act as high capacity parametrized modules by which any affine mixture of features free from the spatial domain could be computed. In the TD processing, such layers as are fully-connected layers in all experiments.

\begin{figure}[t]	
	\begin{center}
		\includegraphics[width=1.0\columnwidth]{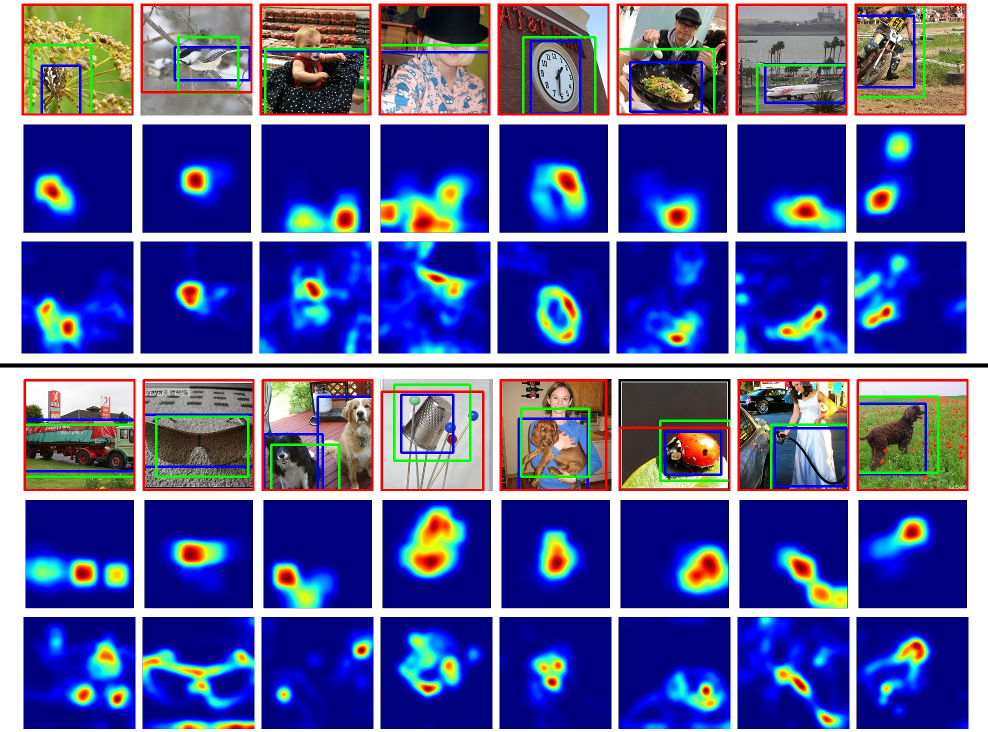}	
	\end{center}	
	\caption{The critical role of the second stage of selection is illustrated using CH visualization. In the top row of each section, images are presented with boxes for the ground truth (blue), full-STNet predictions (green), and second-stage-disabled predictions (red). In the second and third rows of each section, CH maps from the full and partly disabled STNet are given respectively.} 
	\label{fig:phase3}
\end{figure}

\textbf{Context Interference}:
The learned representation of \convnet s strongly relies on the background context over which the category instances are superimposed for the category label prediction of the input image. This is expected since the learning algorithm does not impose any form of spatial regularization during the training phase. Fig. \ref{fig:phase3} depicts the results of the experiment in which we purposefully deactivated the second stage of the selection process. 
Furthermore, among the set of winners at the end of the first stage, the one with the highest PS activity is remained in the set and the rest are excluded. In this way, there is always one winner at each FC layer. 
Deactivating the second stage on the convolutional layers deteriorates the capability of STNet to sharply highlight the salient regions relevant to objects in the TD processing stream. The results implies that the learned representation heavily relies on the features collected across the entire image regardless of the ground truth. The SC mode of the second stage helps STNet to visualize the coherent and sharply localized confident regions.
The CH visualization demonstrates the essential role of the second stage to deal with the redundant and disturbing context noise for the localization task.

\begin{figure}[t]	
	\begin{center}
		\includegraphics[width=1.0\columnwidth]{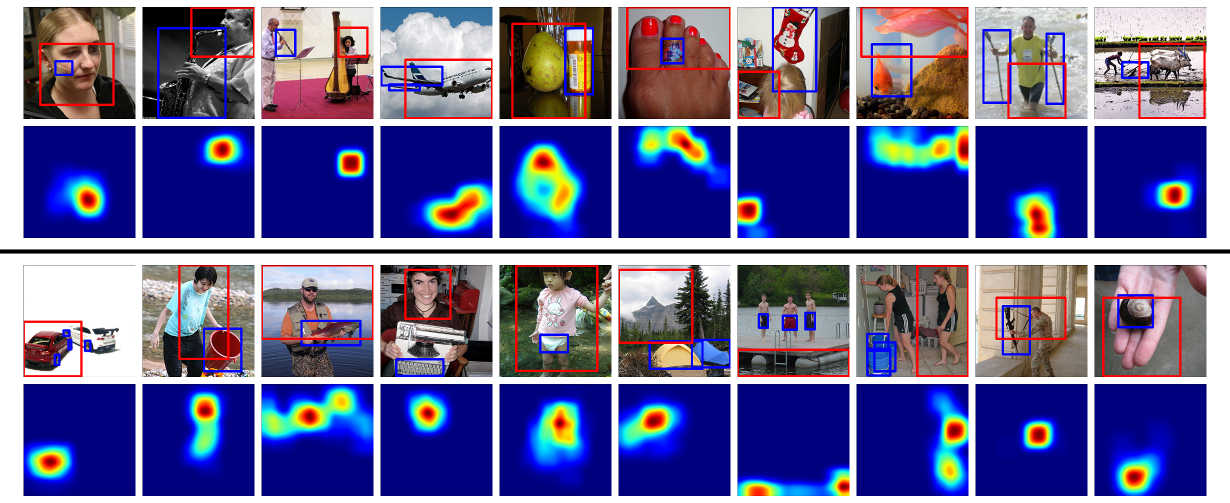}	
	\end{center}	
	\caption{We demonstrate using ST-VGGNet the confident region of the accompanying object that highly correlates with the true object category. In each section, the top row contains images with the ground truth (blue) and predicted (red) boxes. In the bottom rows, CH maps  highlight the most confident salient regions.}
	\label{fig:phase4}
\end{figure}

\textbf{Correlated Accompanying Objects}: 
The other shortcoming of the learned representation emphasized by CH visualization is that the BU processing puts high confidence on the features collected from the regions belonging to correlated accompanying objects. They happen to co-occur extremely frequently with the the ground truth objects in the training set on which \convnet s are pre-trained.
Similar to the previous experiment, the modified version of the first stage for FC layers is used, while the convolutional layers benefit from the original 2-stage selection process. 
Fig. \ref{fig:phase4} reveals how STNet misleadingly localize with the highest confidence the accompanying object that highly correlates with the ground truth object. As soon as the visual representation confidently relates the correlated accompanying object with the true category label, over-selecting for the bounding box prediction will be inevitable.
CAO, in addition to the multi-instance scenario, mostly accounts for the over-selection phenomenon in the localization task. We credit these two sources of over-selection to the pre-trained representation obtained from the backpropagation learning algorithm.


\section{Conclusion}
\label{sec:con}
We proposed an innovative framework consisting of the Bottom-Up and Top-Down streams of information processing for the task of object localization. We formulated the Top-Down processing as a cascading series of local attentive selection processes each consisting of three stages: First inference reduction, second similarity grouping, and third attention signal propagation. 
We demonstrated experimentally the efficiency, power, and speed of STNet to localize objects on the ImageNet dataset supported by the significantly improved quantitative results. 
Class Hypothesis maps are introduced to qualitatively visualize attention-driven class-dependent salient regions. 
Having investigated the difficulties of STNet in object localization, we believe the visual representation of the Bottom-Up stream is one of the shortcomings of this framework. 
The significant role of the selective Top-Down processing in STNet could be foreseen as a promising approach applicable in a similar fashion to other challenging computer vision tasks.

{\small
\bibliographystyle{ieee}
\bibliography{phd}
}

\end{document}